\documentclass[runningheads,a4paper]{llncs}
\usepackage{verbatim}
\usepackage{wrapfig}
\usepackage[table,xcdraw]{xcolor}
\usepackage{multirow}
\usepackage{amssymb}
\usepackage{graphicx}
\usepackage{rotating}
\usepackage{adjustbox}
\usepackage[utf8]{inputenc}
\usepackage{url}
\usepackage{amsmath}
\usepackage{enumitem}
\usepackage{algorithm}
\usepackage[noend]{algpseudocode}
\usepackage{subfig}

\begin{document}

\mainmatter  

\title{Automatically Designing CNN Architectures for Medical Image Segmentation}
\titlerunning{Automatically Designing CNN Architectures}
\author{Aliasghar Mortazi and Ulas Bagci}
\authorrunning{}
\institute{Center for Research in Computer Vision (CRCV), University of Central Florida, Orlando, FL., US. \\ \emph{E-mail: a.mortazi@knights.ucf.edu}}
\toctitle{Lecture Notes in Computer Science}
\tocauthor{Authors' Instructions}
\maketitle
\begin{abstract}
Deep neural network architectures have traditionally been designed and explored with human expertise in a long-lasting trial-and-error process. This process requires huge amount of time, expertise, and resources. To address this tedious problem, we propose a novel algorithm to optimally find hyperparameters of a deep network architecture automatically. We specifically focus on designing neural architectures for medical image segmentation task. Our proposed method is based on a policy gradient reinforcement learning for which the reward function is assigned a segmentation evaluation utility (i.e., dice index).  We show the efficacy of the proposed method with its low computational cost in comparison with the state-of-the-art medical image segmentation networks. We also present a new architecture design, \textit{a densely connected encoder-decoder CNN}, as a strong baseline architecture to apply the proposed hyperparameter search algorithm. We apply the proposed algorithm to each layer of the baseline architectures. As an application, we train the proposed system on cine cardiac MR images from Automated Cardiac Diagnosis Challenge (ACDC) MICCAI 2017. Starting from a baseline segmentation architecture, the resulting network architecture obtains the state-of-the-art results in accuracy without performing any trial-and-error based architecture design approaches or close supervision of the hyperparameters changes.             
\end{abstract}

\section{Introduction}
Deep learning based segmentation algorithms play a key role in medical applications~\cite{unet,segnet,cardiacnet}. However, designing highly accurate and efficient deep segmentation networks is not trivial. It is because manual exploration of high-performance deep networks requires extensive research by close supervision of human expert (from several months to several years) and huge amount of time and resources due to training time of networks. Considering that the choice of architecture and hyperparameters affects the segmentation results, it is extremely important to select the optimal hyperparameters. In this study, we address this pressing problem  by developing a proof of concept optimization algorithm for network architecture design, specifically for medical image segmentation problems.  

Our proposed method is generic and can be applied to any medical image segmentation task. As a proof concept study, we demontrate its efficacy by automatically segmenting heart structures from cardiac magnetic resonance imaging (MRI) scans. Our motivation comes from the fact cardiac MRI plays a significant role in quantification of cardiovascular diseases (CVDs) such that radiologists need to measure the volume of heart and its substructures in association with the cardiac function. This requires a precise segmentation algorithm available in the radiology rooms.  
\begin{figure}[t]
\centering
\includegraphics[width=1\textwidth]{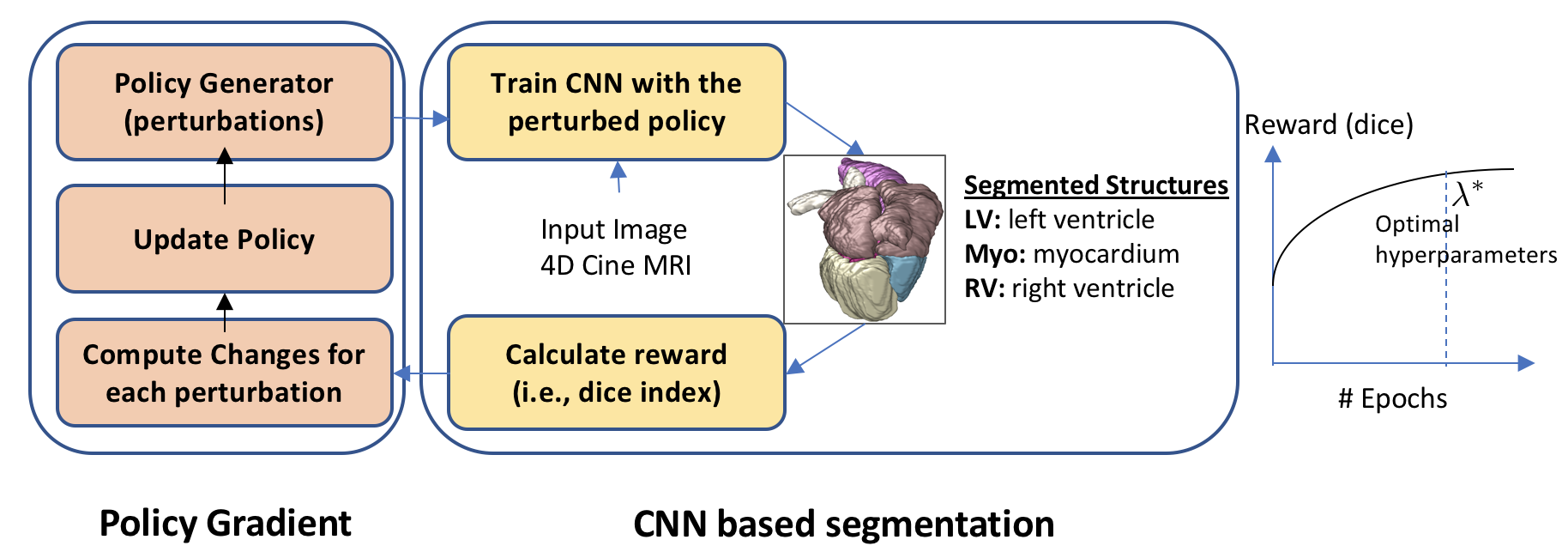}
\caption{Overview of proposed method. First, the policy is initialized randomly and then $P$ perturbation are generated. The network is trained with each perturbation and reward  from each perturbation is calculated. The policy will be updated accordingly and the process will be repeated until no significant changes in the reward. Reward is simply set as dice coefficient for evaluating how good the segmentation is. \label{fig:overall}}
\end{figure}

In recent years, the CNN based deep learning algorithms become the natural choice for medical image segmentation tasks. However, the state-of-the-art CNN-based segmentation methods have very similar fixed network architectures and they all have been designed with a trial-and-error basis. SegNet~\cite{segnet}, CardiacNet~\cite{cardiacnet}, and U-Net~\cite{unet} are some of the notable approaches from the literature. To design such networks, experts have often large number of choices involved in design decisions, and manual search process is significantly guided by intuition. To address this issue, there is a considerable interest recently for designing the network architecture automatically. Reinforcement Learning (RL)~\cite{zoph} and evolutionary based algorithms~\cite{evolutionary} are proposed to search the optimum network hyperparameters. Such methods are computationally expensive  and require a large number of processors (as low as 800 GPUs in Google's network search algorithm~\cite{zoph}) and may not be doable for a widespread and more general use. Instead, in this paper, \textbf{we propose a conceptually simple and very efficient network optimization search algorithm based on a policy gradient (PG) algorithm.} PG is one of the successful algorithms in robotics field~\cite{kohl} for learning system design parameters. Another example is by Zoph and Le~\cite{zoph} where authors used LSTM (long short term memory) to learn the hyperparameters of the CNN and the PG was used to learn the parameters of the LSTM. Learning parameters of LSTM need considerable amount of resources as it is discussed in~\cite{zoph}. Unlike that indirect parameter estimation, in this paper we propose a PG algorithm to directly learn network hyperparameters. Our proposed approach is inspired by~\cite{kohl} and it has been adapted to deep network architecture design for performing image segmentation tasks with high accuracy. In this study, to make the whole system economical to implement for wide range of applications, search space is significantly restricted.

The overview of the proposed method is illustrated in Fig.~\ref{fig:overall}. The hyperparameters of the network are considered as policies to be learned during PG training. To our best of knowledge, this is the first study to find optimum hyperparameters of a given network with policy gradient directly. Moreover, our proposed baseline architecture of densely connected encoder-decoder CNN and the use of \textit{Swish} function as an alternative to \textit{ReLU} are novel and superior to the existing systems. Lastly, our study is the first medical image segmentation work with a fully automated algorithm that discovers the optimal network architecture.

\section{Methods}
\vspace{-0.3cm}
\subsection{Policy Gradient}
Policy gradient is a class of reinforcement learning (RL) algorithms and relied on optimization of parametrized policies with respect to a expected return (reward)~\cite{sutton}. Unlike other RL methods (such as Q-Learning), the PG learns the policy function directly to maximize receiving rewards. In our setting, we consider each hyperparameter of the network as a policy, which can be learned during network training. Assume that we have a policy $\pi_0 = \{ \theta_1, \theta_2, \dots, \theta_N\}$, indicating the hyperparameters of the network, where $\it{N}$ is the number of hyperparameters (dimensions). Our objective is to learn these hyperparameters (i.e., policies) by maximizing a receiving reward. In segmentation task, this reward can be anything measuring the goodness of segmentations such as dice index and Hausdorff distances. Once we randomly initialize hyperparameters, we generate new policies by randomly perturbing the policies in each dimension. Note that each dimension represents an exploration space for hyperparameters such as filter width, hight, and etc. Let $P(\pi_0)=\{\pi_1, \pi_2, \dots, \pi_p\}$ be $\it{p}$ random perturbation generated near $\pi_0$, represented as $\pi_i=\pi_0+\Delta_i$ for $i\in\{1,2,\dots,p \}$. For each random perturbation, $\Delta_i=\{ \delta^1, \delta^2, \dots, \delta^N\}$, we assume that $\delta^d$ is randomly chosen from $\{-\epsilon^d,0,+\epsilon^d\}$ for every $d\in\{1,2,\dots,N\}$ where epsilon is derivative of a function $\mathbf{y}$ with respect to $\mathbf{x}$ (Later we will define $\mathbf{x}$ and  $\mathbf{y}$ for each dimension in Section~\ref{sec:hyper}). 

The network is trained with these $\it{p}$ generated policies, and reward (segmentation outcome) is obtained for each policy. Finally, the maximal reward (i.e., highest dice coefficient) is determined to set the optimal network architecture hyperparameters accordingly. To estimate the partial derivative of the policy function for each dimension, each perturbation is grouped to non-overlapping categories of negative perturbation, zero perturbation, and positive perturbation:  \textit{$C_-^d$}, \textit{$C_0^d$}, and \textit{$C_+^d$} such that $\pi_i^d\in\{C_-^d, C_0^d, C_+^d\}$. The perturbations are generated to make sure each category has approximately $p/3$ members. Then, the absolute reward for each category is calculated as a mean of all the rewards $Ave^d=\{Ave_-^d, Ave_0^d, Ave_+^d \}$ for each dimension \textit{d}. Based on this average reward, the initial policy is updated accordingly: 
\begin{equation}\label{update}
\pi_{0,new}^d=
\begin{cases}
\pi_0^d-\epsilon^d  \text{ if } \quad Ave_-^d> Ave_0^d \quad \text{and } \quad Ave_-^d> Ave_+^d\\
\pi_0^d+0 \text{ if } \quad Ave_0^d \ge Ave_-^d \quad \text{and } \quad Ave_0^d\ge Ave_+^d \\
\pi_0^d+\epsilon^d  \text{ if } \quad Ave_+^d> Ave_0^d \quad \text{and } \quad Ave_+^d> Ave_-^d
\end{cases}
\end{equation}
The pseudo-code for policy gradient is given in algorithm~\ref{alg:PG}. 
\vspace{-0.4cm}
\begin{algorithm}
\caption{Policy Gradient's algorithm}\label{alg:PG}
\begin{algorithmic}[1]
\State Initialize $\pi_0$ randomly
\For{e=1:epochs} 
\State Generate \textit{p} randomly perturbation of $P(\pi_0)=\{\pi_1, \pi_2, \dots, \pi_p\}$
\For {i=1:p}
\State Train network with policy $\pi_i$
\State{Calculate reward}
\EndFor{}
\For{d=1:N}
\State {$Ave_+^d\gets Average$ rewards for $C_+^d$} 
\State {$Ave_0^d\gets Average$ rewards for $C_0^d$} 
\State {$Ave_-^d\gets Average$ rewards for $C_-^d$} 
\State {\textbf{Update} $\pi_{0,new}^d$ based on Equation 1.}
\EndFor{}
\EndFor{}
\end{algorithmic}
\end{algorithm}
\vspace{-0.8cm}
\subsection{Proposed Base-Architecture for Image Segmentation}
\begin{figure}[t]
\vspace{-0.2cm}
\includegraphics[height=4.5cm, width=1\textwidth]{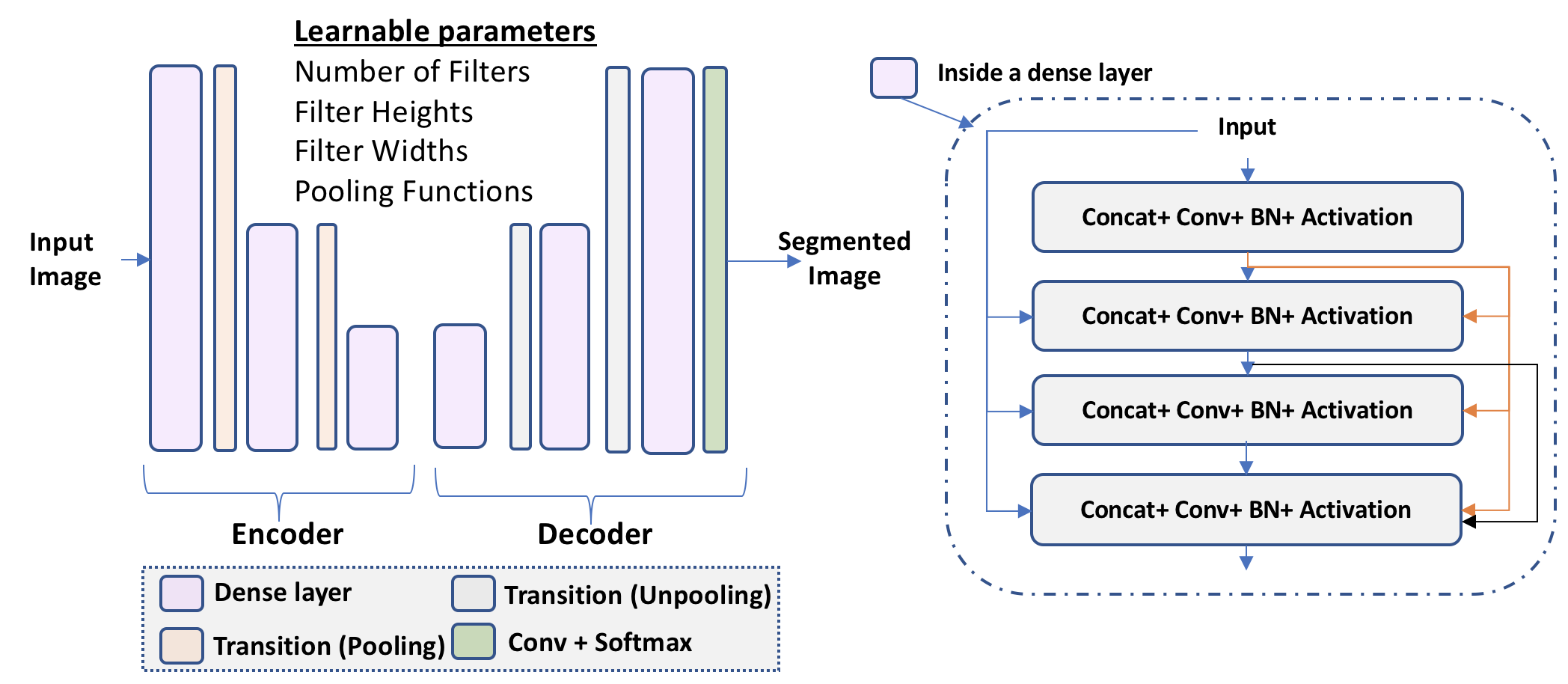}
 \caption{Details of the baseline architecture. We combine encoder-decoder based segmentation network with densely connected architecture as a novel segmentation network, which has less parameters to tune and more accurate. Concat: concatenation, BN: batch normalization, and conv: convolution.\label{fig:CNN-g}}
\vspace{-0.5cm}
\end{figure}

As it has been shown in~\cite{segnet,cardiacnet}, the encoder-decoder architecture is well design deep learning architecture for the segmentation tasks. More recently, the densely connected CNN~\cite{denselyCNN} has been shown that connecting different layers lead into more accurate results for detection problem. Based on this recent evidence, a densely connected encoder-decoder is proposed herein as a new CNN architecture and we use this as our baseline architecture to optimize. The proposed baseline architecture is illustrated in Figure~\ref{fig:CNN-g}. Dense blocks consist of four layers, each layer includes convolution operation following by batch normalization operation (BN) and \textit{Swish} activation function~\cite{swish} (unlike commonly used ReLU). Also, a concatenation operation is conducted for combining the feature maps (through direction (axis) of the channels) for the last three layers. In other words, if the input to $\it{l^{th}}$ layer is $\mathbf{X}_{\it{l}}$, then the output of $\it{l^{th}}$ layer can be represented as:
\begin{equation}\label{layer_eq}
F(\mathbf{X}_{\it{l}})=Conv(BN(Swish(\mathbf{X}_{\it{l}}))), 
\end{equation}
where $Swish(x)=xSigmoid(\beta x)$ and as it is discussed in ~\cite{swish}, the \textbf{Swish} was shown to be more powerful than ReLu since parameter $\beta$ can be learned during training to control the interpolation between linear function ($\beta = 0$) and ReLu function ($\beta \approx \infty$). Since we are doing concatenation before each layer (except the first one), so the output of each layer can be calculated only by considering the input and output of first layer as: 
\begin{equation}\label{concat}
	F(\mathbf{X}_{\it{l}})=F(\displaystyle \underset{{\it{l}'=0}}{\overset{{\it{l}'=\it{l}-1}}{\|}} F(\mathbf{X}_{\it{l}'})) \qquad for \quad\it{l}\ge1 \quad and \quad \it{l}=\{1,2,\dots,\it{L}\},
\end{equation}
where $\|$ is the concatenation operation. For initialization $F(\mathbf{X}_{\it{-1}})$ and $F(\mathbf{X}_{\it{0}})$ are considered as $\phi$ and $\mathbf{X}_{\it{1}}$, respectively, which $\phi$ is an empty set and there are $\it{L}$ layers inside each block.


The decoder part of the CNN consists of three dense blocks and two transition layers. The decoder transition layers can be \textit{average pooling} or \textit{max pooling} and decrease the size of the image by half. In the encoder part, we have same architecture as decoder part except that the transition layers are bilinear interpolation (i.e., unpooling). Each of the decoder transition doubles the size of the feature maps and at the end of this part, we obtain features maps as the same size as input images. Finally, the output of the decoder is passed through a convolution and softmax to produce the probability map. \textit{Adam optimizer} with a learning rate of 0.0001 is selected for training and \textit{Cross Entropy} is used as a loss function. The other hyperparameters of network such as number of filters, filter heights, and widths for each layer are discussed in next section. 

\vspace{-0.2cm}
\subsection{Learnable Hyperparameters}\label{sec:hyper}
\vspace{-0.1cm}
Following hyperparameters are learned automatically with our proposed architecture search algorithm: number of filters, filter height, and filter width for each layer. Additionally, type of pooling layer was considered as learnable hyperparameters in our setting. Totally, there are 76 parameters (N) to be learned: 3 parameters (filter size, height, and weight) for each of 25 layers (last layer has fixed number of filters), and 2 additional hyperparameters (average or max pooling) for down-sampling layers. More specifically: 
\begin{itemize}
\item \textbf{Number of filters:} The number of filters (NF) for each layer is chosen from function $y_{NF}=16x_{NF}+16$ which $x_{NF}=\{1,2,\dots, 12\}$. 
\item \textbf{Filter height:} The filter height (FH) for each layer is chosen from function $y_{FH}=2x_{FH}+1$ which $x_{FH}=\{0,1,\dots,5\}$.
\item \textbf{Filter width:} The filter width (FW) for each layer is chosen from function $y_{FW}=2x_{FW}+1$ which $x_{FW}=\{0,1,\dots,5\}$.
\item \textbf{Pooling functions:} The pooling layer is chosen from function $y_{poolimg}=x_{pooling}$ which $x_{pooling}=\{0,1\}$ which '0' represents max pooling and '1' represents average pooling. 
\end{itemize}
The number of generated perturbation $p$ is considered as 42 (experimentally) and in order to decrease the computational cost, each network is trained for 50 epochs, which is adequate to determine a stable reward for the network. The average of dice index for the last 5 epochs on the held-out validation set is considered as reward for the reinforcement learning.


\section{Experiments and Results}
\vspace{-0.2cm}
\textbf{Dataset:} We used Automatic Cardiac Diagnosis Challenge (ACDC-MICCAI Workshop 2017) data set for evaluation of the proposed system. This dataset is composed of 150 cine-MR images including 30 normal cases, 30 patients with myocardium infarction, 30 patients with dilated cardiomyopathy, 30 patients with hypertrophic cardiomyopathy, and 30 patients with abnormal right ventricle (RV). While 100 cine-MR images were used for training, the remaining 50 images were used for testing. We have applied data augmentation methods, as described in Table 1, prior to training. The MR images were obtained using two MRI scanners of different magnetic strengths (1.5T and 3.0T). Cine MR images were acquired in breath hold (and gating) with a SSFP sequence in short axis. A series of short axis slices cover the LV from the base to the apex, with a thickness of 5 mm (or sometimes 8 mm) and sometimes an inter-slice gap of 5 mm. The spatial resolution goes from 1.37 to 1.68 $mm^2/pixel$ and 28 to 40 volumes cover completely or partially the cardiac cycle.

\begin{wraptable}{r}{0.42\textwidth}
\vspace{-0.5cm}
\centering
\caption{Data augmentation}
\vspace{-0.3cm}
\begin{adjustbox}{max width=0.5\textwidth}
\label{my-label}
\scriptsize
\begin{tabular}{|c|c|l|}
\hline
\multicolumn{3}{|c|}{\cellcolor[HTML]{C0C0C0}\textbf{Data augmentation}}        \\ \hline
\textbf{Methods}                     & \multicolumn{2}{c|}{\textbf{Parameters}} \\ \hline
{Rotation}                    & \multicolumn{2}{c|}{\textbf{$k\times45, k\, \varepsilon [-1,1]$}}        \\ \hline
{Scale}                       & \multicolumn{2}{c|}{\textbf{$\varepsilon [1.3,1.5]$}}        \\ \hline
\multicolumn{3}{|c|}{\cellcolor[HTML]{C0C0C0}\textbf{Training Images}}          \\ \hline
\textbf{\# of Images}                & \multicolumn{2}{c|}{\textbf{Image size}} \\ \hline
{8470}                        & \multicolumn{2}{c|}{{$200\times 200$}}    \\ \hline
\end{tabular}
\end{adjustbox}
\vspace{-0.5 cm}
\end{wraptable}


\textbf{Implementation details:} We calculated dice index (DI) and Hausdorff distance (HD) to evaluate segmentation accuracy (blind evaluation through challenge web page on the test data). The quantitative results for LV (left ventricle), RV, and Myo (myocardium) as well as mean accuracy (Ave.) are shown in Table~\ref{dice}. Twenty images were randomly selected out of the 100 training images as validation set. After finding optimized hyperparameters, the network with learned hyperparameters was trained fully with the augmented data. The augmentation was done with in-plane rotation and scaling (Table 1). The number of images increased by factor of five after augmentation.  

\textbf{Post-Processing:} To have a fair comparison with other segmentation methods, which often use post-processing for improving their segmentation results, we also applied post-processing to refine (improve) the overall segmentation results of all compared methods. We presented our results with and without post-processing in Table 2. Briefly, a 3D fully connected Conditional Random Field (CRF) method was used to refine the segmentation results, taking only a few additional milliseconds. The output probability map of the CNN is used as unary potential and a Gaussian function was used as pairwise potential. Finally, a connected component analysis was applied for further removal of isolated points.

\begin{figure}[t]
\vspace{-0.5cm}
\centering
\includegraphics[width=1\textwidth]{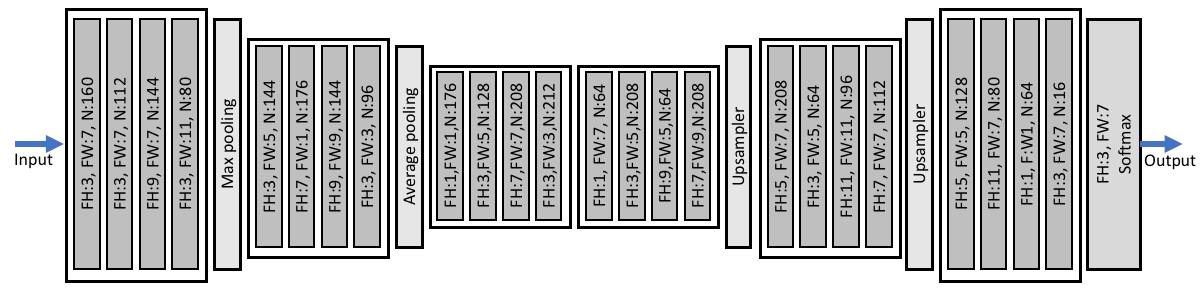}
\caption{Details of the optimally learned architecture by the proposed method. Note that connections among layers inside of each block are same as dense layers.\label{fig:learned}}
\vspace{-0.5cm}
\end{figure}

\textbf{Comparison to other methods:} The performances of the proposed segmentation algorithm in comparison with state-of-the-art methods are summarized in Table~\ref{dice}. The DenseCNN (with ~\textit{ReLu} and with ~\textit{Swish}) is the densely connected encoder-decoder CNN designed by experts, and its use in segmentation tasks recently appeared in some few applications, but never used for cardiac segmentation before. Filter sizes were all set to $3\times3$ in DenseCNN and growth rates were considered as 32, 64, 128, 128, 64, and 32 for each block from beginning to the end of the network, respectively. Also, the average pooling is chosen as the pooling layer. These values were all found after trial-error and empirical experiences, guided by expert opinions as dominant in this field. The 2D U-Net, as one of the state of the arts, is the original implementation of the U-Net architecture proposed by Ronneberger et al. in~\cite{unet} was used for comparison too. Although we apply our algorithm into 2D setting for efficiency purpose, one can apply it to 3D architectures once memory and other hardware constraints are solved. The details of the learned architecture with the proposed method is shown in Fig.~\ref{fig:learned}.

We obtained the final architecture design in 10 days of continuous training of a workstation with 15 GPUs (Titan X). Unlike the common CNN architecture designs (expert approach), which requires months or even years of trial-and-error and experience guided search, the proposed search algorithm found optimal (or near-optimal) segmentation results compared to the state of the art segmentation architectures within days. 

\begin{table}[]
\vspace{-0.9cm}
\centering
\scriptsize
\caption{DI and HD for all methods and substructures.}
\label{dice}
\begin{tabular}{|c|c|c|c|c|c|c|}
\hline
\rowcolor[HTML]{EFEFEF} 
\multicolumn{2}{|c|}{\cellcolor[HTML]{EFEFEF}Methods} & \begin{tabular}[c]{@{}c@{}}2D-UNET\end{tabular} & \begin{tabular}[c]{@{}c@{}}DenseCNN\\ (ReLU)\end{tabular} & \begin{tabular}[c]{@{}c@{}}DenseCNN \end{tabular} & Proposed & Proposed+CRF \\ \hline
 & LV & 0.904 & 0.913 & 0.922 & 0.921 & ~\textbf{0.928} \\ \cline{2-7} 
 & RV & 0.868 & 0.826 & 0.834 & 0.857 & ~\textbf{0.868} \\ \cline{2-7} 
 & MYO & 0.847 & 0.832 & 0.845 & 0.838 & ~\textbf{0.849} \\ \cline{2-7} 
\multirow{-4}{*}{DI} & Ave. & 0.873 & 0.857 & 0.867 & 0.872 & ~\textbf{0.88}2 \\ \hline
 & LV & 9.670 & 9.15 & 8.937 & 8.99 & ~\textbf{8.90} \\ \cline{2-7} 
 & RV & 14.37 & 16.35 & 16.31 & 14.27 & ~\textbf{14.13} \\ \cline{2-7} 
 & MYO & 12.13 & 11.32 & 11.28 & 10.70 & ~\textbf{10.66} \\ \cline{2-7} 
\multirow{-4}{*}{\begin{tabular}[c]{@{}c@{}}HD\\ (mm)\end{tabular}} & Ave. & 12.06 & 12.27 & 13.02 & 11.32 & ~\textbf{11.23} \\ \hline
\end{tabular}
\vspace{-0.5cm}
\end{table}



\vspace{-0.5cm}
\section{Discussions and Conclusion}
\vspace{-0.2cm}
We proposed a new deep network architecture to automatically segment cardiac cine MR images. Our architecture design was fully automatic and based on policy gradient reinforcement learning. After baseline network was structured based on densely connected encoder-decoder network, the policy gradient algorithm automatically searched the hyperparameters of this network, achieving the state of the art results. \textbf{Note that our hypothesis was to show that it was possible to design CNN automatically for medical image segmentation with similar or better performance in accuracy, and much better in efficiency. It is because expert-design networks require extensive trial-and-error experiments and may take even years to design.} Our study has opened a new venue for designing a segmentation  engine within a short period of time. Our study has some limitations  due to its proof of concept nature. One interesting way to extend the proposed model will be to learn hyperparameters conditionally in each layer (unlike independent assumption of the layers). With the availability of more hardware sources, one may explore many more hyperparameters, such as ability to put more layers than basic model, defining skip-connections, and exploring different activation functions instead of ReLU and other default ones. One may also avoid increasing search space and still perform a good architecture design automatically by choosing the base-architecture more powerful ones such as the \textbf{SegCaps} (i.e., segmentation capsules)~\cite{segcaps}.




\bibliographystyle{IEEEbib}
\vspace{-0.1cm}
\bibliography{refs}
\end{document}